\documentclass[sigconf]{acmart}
\copyrightyear{2022}
\acmYear{2022}
\setcopyright{acmcopyright}
\acmConference[KDD '22] {Proceedings of the 28th ACM SIGKDD Conference on Knowledge Discovery and Data Mining}{August 14--18, 2022}{Washington, DC, USA.}
\acmBooktitle{Proceedings of the 28th ACM SIGKDD Conference on Knowledge Discovery and Data Mining (KDD '22), August 14--18, 2022, Washington, DC, USA}
\acmPrice{15.00}
\acmISBN{978-1-4503-9385-0/22/08}
\acmDOI{10.1145/3534678.3539260}

\newcommand{\OURS}{\textsc{LTP}\xspace}
\newcommand{\topk}{top-$k$\xspace}

\usepackage{siunitx}

\newcommand{\commentout}[1]{}

\definecolor{light-gray}{gray}{0.80}

\newcommand\aref{Algorithm~\ref}

\newcommand\eref{Eq.~\ref}
\newcommand\fref{Figure~\ref}
\newcommand\tref{Table~\ref}
\newcommand\sref{Section~\ref}

\newcommand\ha{ \rowcolor{orange!0}}

\newcommand\hc{ \rowcolor{orange!10}}

\usepackage{amsthm}

\usepackage{amsthm}

\newtheorem{mytheorem}{Theorem}
\newtheorem{assumption}[mytheorem]{Assumption}
\newtheorem{lemma}[mytheorem]{Lemma}
\newtheorem{remk}[mytheorem]{Remark}

\usepackage[utf8]{inputenc} 
\usepackage[T1]{fontenc}    
\usepackage{url}            
\usepackage{booktabs}       
\usepackage{amsfonts}       
\usepackage{nicefrac}       
\usepackage{microtype}      
\usepackage{soul}
\usepackage[utf8]{inputenc}
\usepackage{graphicx}
\usepackage{amsmath}
\usepackage{booktabs}
\usepackage{outlines}
\usepackage{adjustbox}

\usepackage{subfig}

\usepackage{microtype}
\usepackage{graphicx}
\usepackage{booktabs} 

\usepackage{amsmath,amsfonts,bm}

\usepackage{xcolor}
\usepackage{colortbl}

\def\eqref#1{equation~\ref{#1}}

\def\rmA{{\mathbf{A}}}

\def\rmW{{\mathbf{W}}}

\DeclareMathAlphabet{\mathsfit}{\encodingdefault}{\sfdefault}{m}{sl}
\SetMathAlphabet{\mathsfit}{bold}{\encodingdefault}{\sfdefault}{bx}{n}

\usepackage{multirow}
\usepackage{paralist}

\usepackage{booktabs} 
\usepackage{multicol}
\usepackage{diagbox}

\usepackage[ruled,noend,algo2e]{algorithm2e}

\SetCommentSty{mycommfont}

\usepackage{here}

\usepackage{tabularx,colortbl,xcolor}
\usepackage{multirow}
\usepackage[normalem]{ulem}
\useunder{\uline}{\ul}{}

\usepackage{enumitem}

\usepackage{xparse}

\DeclareGraphicsExtensions{.pdf,.png}

\usepackage{algorithm} 
\usepackage{algorithmicx}
\usepackage{algpseudocode}

\begin{document}

\title{Learned Token Pruning for Transformers}

\author{Sehoon Kim}
\authornote{Equal contribution.}
\email{sehoonkim@berkeley.edu}
\affiliation{
  \institution{University of California, Berkeley}
  \city{Berkeley}
  \state{CA}
  \country{USA}
}

\author{Sheng Shen}
\authornotemark[1]
\email{sheng.s@berkeley.edu}
\affiliation{
  \institution{University of California, Berkeley}
  \city{Berkeley}
  \state{CA}
  \country{USA}
}

\author{David Thorsley}
\authornotemark[1]
\email{d.thorsley@samsung.com}
\affiliation{
  \institution{Samsung Semiconductor, Inc.}
  \city{San Jose}
  \state{CA}
  \country{USA}
}

\author{Amir Gholami}
\authornotemark[1]
\email{amirgh@berkeley.edu}
\affiliation{
  \institution{University of California, Berkeley}
  \city{Berkeley}
  \state{CA}
  \country{USA}
}

\author{Woosuk Kwon}
\email{woosuk.kwon@berkeley.edu}
\affiliation{
  \institution{University of California, Berkeley}
  \city{Berkeley}
  \state{CA}
  \country{USA}
}

\author{Joseph Hassoun}
\email{j.hassoun@samsung.com}
\affiliation{
  \institution{Samsung Semiconductor, Inc.}
  \city{San Jose}
  \state{CA}
  \country{USA}
}

\author{Kurt Keutzer}
\email{keutzer@berkeley.edu}
\affiliation{
  \institution{University of California, Berkeley}
  \city{Berkeley}
  \state{CA}
  \country{USA}
}

\renewcommand{\shortauthors}{Sehoon Kim, Sheng Shen, David Thorsley, Amir Gholami, Woosuk Kwon, Joseph Hassoun, Kurt Keutzer}


\begin{abstract}
Efficient deployment of transformer models in practice is challenging due to their inference cost including memory footprint, latency, and power consumption, which scales quadratically with input sequence length.
To address this, we present a novel token reduction method dubbed Learned Token Pruning (\OURS) 
which adaptively removes unimportant tokens as an input sequence passes through transformer layers.
In particular, \OURS prunes tokens with an attention score below a threshold,
whose value is learned for each layer during training.
Our threshold-based method allows the length of the pruned sequence to vary adaptively based on the input sequence,
and avoids algorithmically expensive operations such as top-$k$ token selection.
We extensively test the performance of \OURS on GLUE and SQuAD tasks and show that our method outperforms the prior state-of-the-art token pruning methods by up to $\sim$2.5\% higher accuracy with the same amount of FLOPs.
In particular, \OURS achieves up to 2.1$\times$ FLOPs reduction with less than 1\% accuracy drop,
which results in up to 1.9$\times$ and 2.0$\times$ throughput improvement on Intel Haswell CPUs and NVIDIA V100 GPUs.
Furthermore, we demonstrate that \OURS is more robust than prior methods to variations in input sequence lengths.
Our code has been developed in PyTorch and open-sourced\footnote{\href{https://github.com/kssteven418/LTP}{https://github.com/kssteven418/LTP}}.
\end{abstract}





\maketitle

\section{\textbf{ Introduction}}
\label{sec:intro}
Transformer-based deep neural network architectures \cite{vaswani2017attention}, such as BERT~\cite{devlin2019bert} and RoBERTa~\cite{liu2019roberta},
achieve state-of-the-art results in Natural Language Processing (NLP) tasks such as sentence classification and question answering.
However, efficiently deploying these models is increasingly challenging due to their large size, the need for real-time inference, and the limited energy, compute, and memory resources available.
The heart of a transformer layer is the multi-head self-attention mechanism, where each token in the input sequence attends to every other token to compute a new representation of the sequence.
Because all tokens attend to each others, the computation complexity is quadratic with respect to the input sequence length; thus the ability to apply transformer models to long input sequences becomes limited.
\begin{figure*}[!ht]
\centering
\subfloat[QQP]{\includegraphics[width=.32 \textwidth]{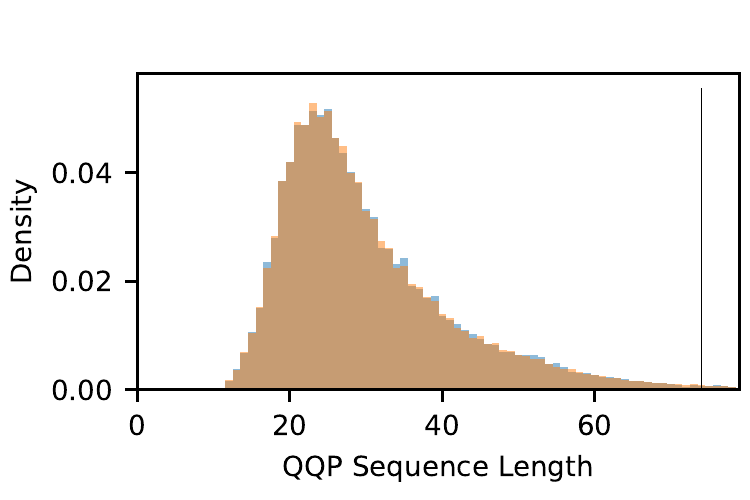}}\quad
\subfloat[SST-2]{\includegraphics[width=.32 \textwidth]{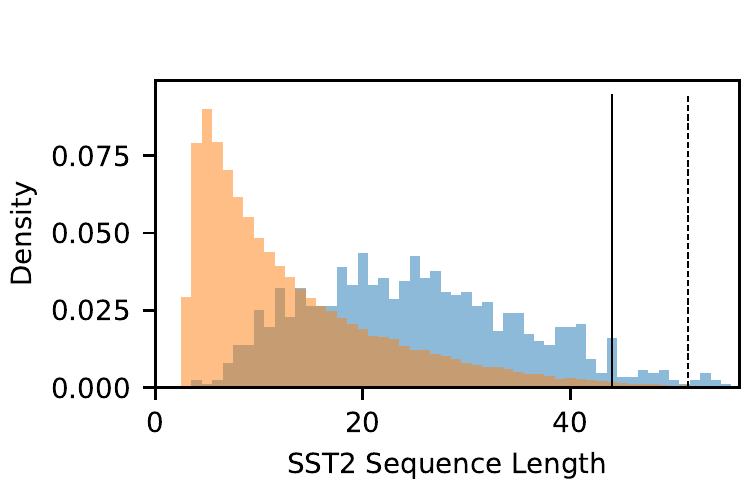}}\quad
\subfloat[STS-B]{\includegraphics[width=.32 \textwidth]{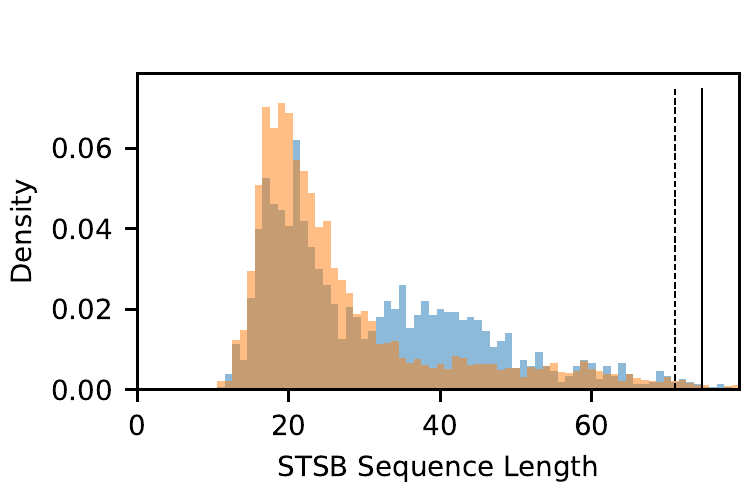}}
\caption{Distributions of processed input sequence lengths from datasets for representative tasks in the GLUE benchmark: (a) QQP (b) SST-2; (c) STS-B. The training set is in orange and the validation set is in blue. The dashed and solid vertical lines indicate the 99th percentile value for the training and validation sets, respectively.}

\label{fig:glue-distributions}
\end{figure*}

Pruning is a popular technique to reduce the size of neural networks and the amount of computation required.
{\it Unstructured} pruning allows arbitrary patterns of sparsification for parameters and feature maps and can, in theory, produce significant computational savings while preserving accuracy.
However, commodity DNN accelerators cannot efficiently exploit unstructured sparsity patterns.
Thus, {\it structured} pruning methods are typically preferred in practice due to their relative ease of deployment to hardware.

Multi-head self-attention provides several possibilities for structured pruning; for example,
head pruning \cite{michel2019sixteen, voita2019analyzing} decreases the size of the model by removing unneeded heads in each transformer layer.
Another orthogonal approach that we consider in this paper
is {\it token pruning}, which reduces computation by progressively
removing unimportant tokens in the sequence during inference.
For NLP tasks such as sentence classification,
token pruning is an attractive approach to consider as it exploits the intuitive observation
that not all tokens (i.e., words) in an input sentence are necessarily required to make a successful inference.

There are two main classes of token pruning methods.
In the first class, methods like PoWER-BERT \cite{goyal20powerbert} and Length-Adaptive Transformer (LAT) \cite{kim2021lat} search for a single token pruning configuration (i.e., sequence length for each layer) for an entire dataset.
In other words, they prune all input sequences to the same length.
However, input sequence lengths can vary greatly within tasks and between training and validation sets as in \fref{fig:glue-distributions}, and thus applying a single pruning configuration to all input sequences can potentially under-prune shorter sequences or over-prune longer sequences.

In the other class, the token pruning method adjusts the  configuration based on the input sequence. SpAtten~\cite{wang2020spatten} uses a pruning configuration proportional to input sentence length; however, it does not adjust the proportion of pruned tokens based on the content of the input sequence. The recently published TR-BERT \cite{ye2021trbert} uses reinforcement learning (RL) to find a policy network that dynamically reduces the number of tokens based on the length and content of the input sequence;
however, it requires additional costly training for convergence of the RL-based method.
Additionally, all of these prior methods rely in part on selecting the $k$ most significant tokens during inference or training.
This selection can be computationally expensive without the development of specialized hardware, such as the \topk engine introduced in SpAtten~\cite{wang2020spatten}.

To this end, we propose a learned \textit{threshold}-based token pruning method which adapts to the length and content of individual examples and avoids the use of \topk operations. In particular, our contributions are as follows:

\begin{itemize}[leftmargin=*]
    \item We propose Learned Token Pruning (\OURS), a threshold-based token pruning method, which only needs a simple threshold operation to detect unimportant tokens.
    In addition, \OURS fully automates the search for optimal pruning configurations 
    by introducing a differentiable soft binarized mask that allows training the correct thresholds for different layers and tasks. 
    (\sref{sec:learned_threshold})
    
    \vspace{1mm}
    \item We apply \OURS to RoBERTa and evaluate its performance on GLUE and SQuAD tasks. 
    We show \OURS achieves up to 2.10$\times$ FLOPs reduction 
    with less than 1\% accuracy degradation,
    which results in up to 1.93$\times$ and 1.97$\times$ throughput improvement on NVIDIA V100 GPU and Intel Haswell CPU, respectively, as compared to the unpruned FP16 baseline.
    We also show that \OURS outperforms SpAtten and LAT in most cases, achieving additional FLOPs reduction for the same drop in accuracy.
    (\sref{sec:evaluation} and \ref{sec:throughput})
    
    \vspace{1mm}
    \item We show that \OURS is highly robust against sentence length variations.
    \OURS exhibits consistently better accuracy over different sentence length distributions, achieving up to 16.4\% accuracy gap from LAT.
    (\sref{sec:robustness_eval})

\end{itemize}

\begin{figure*}[!t]
\centering
    \includegraphics[width=0.99\textwidth]{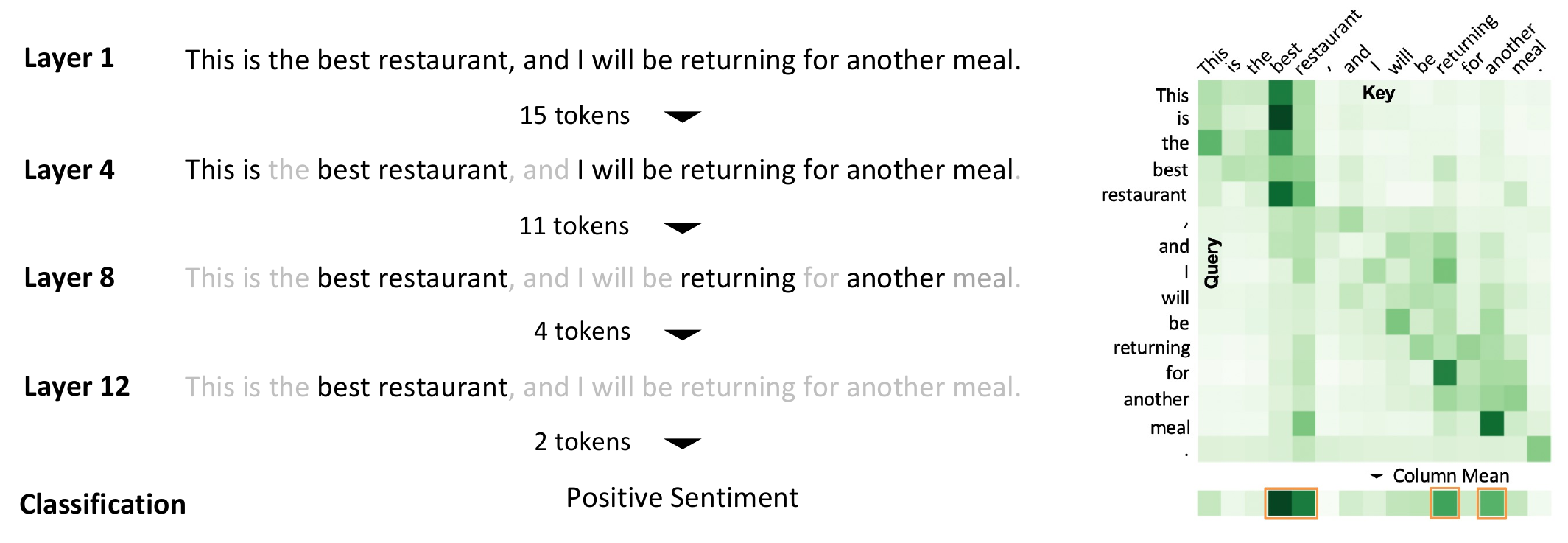}
\caption{
(Left) Schematic of token pruning for a sentiment analysis task. 
Unimportant tokens are pruned as the input sequence passes through the layers. 
(Right) An example of attention probability in a single head where a more important token receives more attention from other tokens. 
Thus each token's importance score is computed by taking the average attention probability it receives, which is computed by taking the column mean of the attention probability.
}
\label{fig:sentence-examples}
\end{figure*}

\section{\textbf{Related Work}}
\label{sec:relatedwork}
\vspace{1mm}

\subsection{Efficient Transformers}
Multiple different approaches have been proposed to improve speed and diminish memory footprint of transformers.
These can be broadly categorized as follows:
(i)  efficient architecture design~\cite{lan2019albert,child2019generating,kitaev2020reformer,wang2020linformer,iandola2020squeezebert,vyas2020fast,tay2020sparse,katharopoulos2020transformers,zaheer2020big,roy2021efficient}; 
(ii) knowledge distillation~\cite{sun2020mobilebert,jiao2019tinybert,tang2019distilling,sanh2019distilbert,sun2019patient}; 
(iii) quantization~\cite{bhandare2019efficient,zafrir2019q8bert,shen2020q,fan2020training,zadeh2020gobo,zhang2020ternarybert,bai2020binarybert,kim2021bert};
and (iv) pruning. 
Here, we focus only on pruning and briefly discuss the
related work.

\subsection{Transformer Pruning}
Pruning methods can be categorized into unstructured pruning and structured pruning. 
For unstructured pruning, the lottery-ticket hypothesis~\cite{frankle2018lottery} has been explored for transformers in \cite{prasanna2020bert,chen2020lottery}. 
Recently, \cite{zhao2020masking} leverages pruning as an effective way to fine-tune transformers on downstream tasks. 
\cite{sanh2020movement} proposes movement pruning, which achieves significant performance improvements versus prior magnitude-based methods by considering the weights modification during fine-tuning. 
However, it is often quite difficult to efficiently deploy unstructured sparsity on commodity neural accelerators for meaningful speedup.

For this reason, a number of structured pruning methods have been introduced to remove structured sets of parameters.
\cite{michel2019sixteen, voita2019analyzing} drop attention heads in multi-head attention layers, 
and \cite{sajjad2020effect, fan2019reducing} prunes entire transformer layers.
\cite{wang2019structured} structurally prunes weight matrices via low-rank factorization, and \cite{khetan2020schubert, lin2020pruning} attempt to jointly prune attention heads and filters of weight matrices.
\cite{liu2021ebert,hou2020dynabert} dynamically determines structured pruning ratios during inference.
Recent block pruning schemes chunk weight matrices into multiple blocks and prune them based 
on group Lasso optimization~\cite{li2020efficient}, adaptive regularization~\cite{yao2021mlpruning}, and movement pruning~\cite{lagunas2021block}.
All of these methods correspond to \textit{weight pruning}, where model parameters (i.e., weights) are pruned.

Recently, there has been work on pruning input sentences to transformers, rather than model parameters.
This is referred to as \textit{token pruning}, where less important tokens are progressively removed during inference.
PoWER-BERT~\cite{goyal20powerbert}, one of the earliest works, proposes to directly learn token pruning configurations.
LAT~\cite{kim2021lat} extends this idea by introducing LengthDrop, a procedure in which a model is trained with different token pruning configurations, followed by an evolutionary search. 
This method has an advantage that the former training procedure need not be repeated for
different pruning ratios of the same model.
While these methods have shown a large computation reduction on various NLP downstream tasks, 
they fix a single token pruning configuration for the entire dataset.
That is, they prune all input sequences to the same length.
However, as shown in~\fref{fig:glue-distributions}, input sequence lengths vary greatly within a task.
As a consequence, fixing a single pruning configuration can under-prune shorter sequences so as to retain sufficient tokens for processing longer sequences or, conversely, over-prune longer sequences to remove sufficient tokens to efficiently process shorter sequences.
More importantly, a single pruning configuration lacks robustness against input sequence length variations, where input sentences at inference time are longer than those in the training dataset~\cite{press2021train}.

In contrast, SpAtten~\cite{wang2020spatten} circumvents this issue by assigning a pruning configuration proportional to the input sequence length.
While this allows pruning more tokens from longer sequences and fewer tokens from shorter ones,
it is not adaptive to individual input sequences as it assigns the same configuration to all sequences with the same length regardless of their contents.
In addition, the pruning configurations are determined heuristically and thus can result in a suboptimal solution.
Recently, TR-BERT~\cite{ye2021trbert} proposes to learn a RL policy network to apply adaptive pruning configurations for each input sequence.
However, as noted by the authors, the problem has a large search spaces which can be hard for RL to solve.
This issue is mitigated by heuristics involving imitation learning and sampling of action sequences, which significantly increases the cost of training.
Importantly, all of the aforementioned token pruning methods depend partially or entirely on \topk operation for selecting the $k$ most important tokens during inference or training.
This operation can be costly without specialized hardware support, as discussed in~\cite{wang2020spatten}.
\OURS, on the other hand, is based on a fully learnable threshold-based pruning strategy.
Therefore, it is (i) adaptive to both input length and content,
(ii) robust to sentence length variations, (iii) computationally efficient, and (iv) easy to deploy.

\vspace{2mm}
\section{\textbf{Methodology}}
\vspace{1mm}
\label{sec:methodology}

\begin{figure*}
\centering{
\centerline{
\centering
  \includegraphics[width=\textwidth]{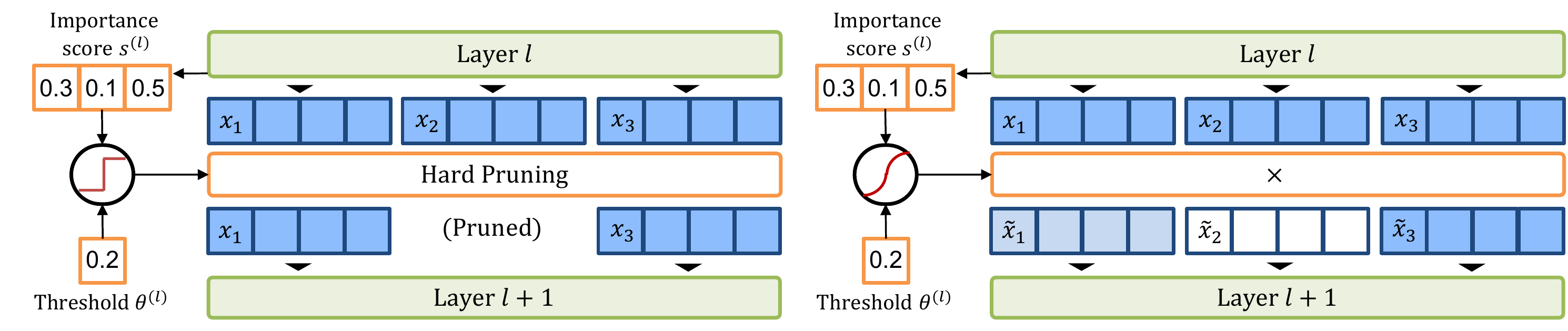}
  }
  \caption{ Different pruning strategies for threshold-based token pruning methods. 
  (Left) Hard pruning uses a binary hard mask to select tokens to be pruned. 
  (Right) Soft pruning replaces the binary mask with a differentiable soft mask.
  }
  \label{fig:overview}
  }
 \end{figure*}

\subsection{Background}
\label{sec:background}

BERT~\cite{devlin2019bert} consists of multiple transformer encoder layers~\cite{vaswani2017attention} stacked up together. 
A basic transformer encoder layer consists of a multi-head attention (MHA) block followed by a point-wise feed-forward (FFN) block, with residual connections around each. 
Specifically, an MHA consists of $N_h$ independently parameterized heads. 
An attention head $h$ in layer $l$ is parameterized by $ \rmW_k^{(h,l)},~\rmW_q^{(h,l)},~\rmW_v^{(h,l)} \in \mathbb{R}^{d_h \times d} $, $ \rmW_o^{(h,l)} \in \mathbb{R}^{d \times d_h} $, 
where $d_h$ is typically set to $d/N_h$ and $d$ is the feature dimension. 
We drop the superscript $l$ for simplicity in the following formula. 
The MHA 
measures the pairwise importance of each token on every other token in the input:
\begin{equation}
\label{eq:mhatt}
\text{MHA}(\text{x}) = 
\sum_{h=1}^{N_h} \text{Att}_{\rmW_{k,q,v,o}^{(h)}}(\text{x}),
\end{equation}
where $\text{x} \in \mathbb{R}^{d \times n}$ is the input sequence with the sequence length $n$, and $\text{Att}_{\rmW_{k,q,v,o}}$ is:
\begin{align}
    \text{Att}_{\rmW_{k,q,v,o}}(\text{x})&=\rmW_{o} \sum_{i=1}^n \rmW_{v} \text{x}_i\text{softmax}(\frac{\text{x}^T\rmW_{q}^T\rmW_{k}\text{x}_i}{\sqrt{d}}) ,\\
\label{eq:ln_attn}
    \textbf{x}_{\text{MHA}} &= \text{LN}\big(\text{Att}_{\rmW_{k,q,v,o}}(\text{x}) + \text{x}\big),
\end{align}
where~\eref{eq:ln_attn} is the residual connection and the follow up LayerNorm (LN).
The output of the MHA is then fed into the FFN block which applies two feed-forward layers to this input:
\begin{align}
\text{FFN}(\text{x}_\mathrm{{MHA}}) &= \sigma\big(\rmW_2(\rmW_1\text{x}_\mathrm{{MHA}} + b_1)\big) + b_2, \\
\textbf{x}_\mathrm{{out}} &= \text{LN}\big(\text{FFN}(\text{x}_\mathrm{{MHA}}) + \text{x}_\mathrm{{MHA}}\big),
\end{align}
where $\rmW_1, \rmW_2, b_1$ and $b_2$ are the FFN parameters, and $\sigma$ is the activation function (typically GELU for BERT).

\subsection{Threshold Token Pruning}
\label{sec:threshold_token_pruning}

Let us denote the \textit{attention probability} of head $h$ between token x$_i$ and x$_j$ as $\rmA^{(h, l)}$:
\begin{equation}
\rmA^{(h, l)}(\text{x}_i, \text{x}_j) = \text{softmax}(\frac{\text{x}^T\rmW_{q}^T\rmW_{k}\text{x}}{ \sqrt{d}})_{(i,j)} \in \mathbb{R}. 
\end{equation}
The cost of computational complexity for computing the attention matrix is
$\mathcal{O}(d^2n + n^2d)$, which quadratically scales with sequence length.
As such, the attention operation becomes a bottleneck when applied to
long sequences. To address this, we apply \emph{token pruning}
which removes unimportant tokens as the input passes through the transformer layers to reduce the sequence length $n$ for later blocks.
This is schematically shown in~\fref{fig:sentence-examples} (Left).

For token pruning, we must define a metric to determine unimportant tokens. Following~\cite{goyal20powerbert, wang2020spatten, kim2021lat},
we define the \textit{importance score} of token x$_i$ in layer $l$ as:
\begin{equation}
\label{eq:importance_score}
s^{(l)}(\text{x}_i) = \frac{1}{N_h} \frac{1}{n} \sum_{h=1}^{N_h} \sum_{j=1}^{n} \rmA^{(h, l)}(\text{x}_i, \text{x}_j). 
\end{equation} 
Intuitively, the attention probability $\rmA^{(h, l)}(\text{x}_i, \text{x}_j)$ is interpreted as the normalized amount that all the other tokens x$_j$ attend to token x$_i$.
Token x$_i$ is thus considered \textit{important} if it receives more attention from all tokens across all heads, which directly leads us to~\eqref{eq:importance_score}.
The procedure for computing importance scores from attention probabilities is illustrated in \fref{fig:sentence-examples} (Right). 

In~\cite{goyal20powerbert, wang2020spatten, kim2021lat}, tokens are ranked by 
importance score and pruned using a \topk selection strategy. 
Specially, token x$_i$ is pruned at layer $l$ if its important score $s^{(l)}(\text{x}_i)$
is smaller than the $k$-largest values of the important score from all the tokens.  
However, finding the $k$-largest values of the importance score is computationally
inefficient without specialized hardware~\cite{wang2020spatten};
we provide empirical results showing this in~\sref{sec:computation_efficiency}.
Instead, we introduce a new \textit{threshold-based} token pruning approach in which
a token is pruned only if its importance score is below a threshold denoted by
$\theta^{(l)} \in \mathbb{R}$. Specifically, we define a pruning strategy by imposing a binary mask $M^{(l)}(\cdot): \{ 1, \dots, n \} \to \{0, 1\}$ 
which indicates whether a token should be kept or pruned: 
\begin{equation}
\label{eq:masking_operator}
M^{(l)}(\text{x}_i) =
    \begin{cases}
    1~~&\text{if }s^{(l)}(\text{x}_i) > \theta^{(l)},\\
    0~~&\text{otherwise}.
    \end{cases}
\end{equation}

Note that this operation only requires a simple
comparison operator without any expensive \topk calculation.
Once a token is pruned, it is excluded from calculations in all succeeding layers, thereby gradually reducing computation complexity towards the output layers.

\subsection{Learnable Threshold for Token Pruning}
~\label{sec:learned_threshold}
\vspace{-3mm}

A key concern with the method above is how to determine the threshold values for each layer.
Not only do threshold values change for different layers, they also vary between different tasks.
We address this by making the 
thresholds (i.e., $\theta$ in~\eref{eq:masking_operator}) learnable. 
However, there are several challenges to consider.
First, due to the binary nature of $M$ there is no gradient flow for pruned tokens.
Second, the $M$ operator is non-differentiable which prevents gradient flow into the thresholds.
To address these challenges, we use a \textit{soft} pruning scheme that simulates the original \textit{hard} pruning
while still propagating gradients to the thresholds as shown in~\fref{fig:overview}. 

\vspace{2mm}
\textbf{Soft Pruning Scheme.} In the soft pruning scheme, we replace the non-differentiable mask $M^{(l)}$ with a differentiable soft mask using the sigmoid operation $\sigma$:
\begin{align}
    \tilde{M}^{(l)}(\text{x}_i) &= \sigma \left( \frac{s^{(l)}(\text{x}_i) - \theta^{(l)}}{T}\right),
\end{align}
where $T$ is temperature, and $\theta^{(l)}$ is the learnable threshold value for layer $l$.
With sufficiently small temperature $T$, $\tilde{M}^{(l)}(\text{x}_i)$ will closely approximate the hard masking $M^{(l)}(\text{x}_i)$
in~\eref{eq:masking_operator}.
In addition, instead of selecting tokens to be pruned or kept based on
the hard mask of~\eref{eq:masking_operator}, we multiply the soft mask to the output activation of layer $l$. That is,
\begin{align}
    \tilde{\text{x}}_\mathrm{{out}}^{(l)} &= 
    \tilde{M}^{(l)}(\text{x}^{(l)}) \cdot \text{x}_\mathrm{{out}}^{(l)} \\
    &= \tilde{M}^{(l)}(\text{x}^{(l)}) \cdot \mathrm{LN}
    (\text{FFN}(\text{x}_\mathrm{{MHA}}^{(l)}) + \text{x}_\mathrm{{MHA}}^{(l)}),
\end{align}
where $\text{x}_\mathrm{{MHA}}^{(l)}$ is the output activation of MHA in layer $l$.
If the importance score of token x$_i$ is below the threshold by a large margin, its layer output activation nears zero and thus it has little impact on the succeeding layer. 
Also, because the token gets a zero importance score in the succeeding layer, i.e., $s^{(l+1)}(\text{x}_i) = 0$, it is likely to be pruned again.
Therefore, the soft pruning scheme is nearly identical in behavior to hard pruning, yet its differentiable form
allows for backpropagation and gradient-based optimizations to make $\theta$ learnable. 
After (i) jointly training model parameters and thresholds on downstream tasks with the soft pruning scheme,
(ii) we fix the thresholds and binarize the soft mask, and 
(iii) perform a follow-up fine-tuning of the model parameters.
The pseudo-code for this three-step algorithm is given in~\aref{alg:lttp}.
Intuitively, the magnitude of gradient $d\tilde{M}^{(l)}(\text{x}_i)/d\theta^{(l)}$ is maximized when the importance score $s^{(l)}(\text{x}_i)$ is close enough to the threshold $\theta^{(l)}$ and becomes near zero elsewhere.
Therefore, the threshold can be trained only based on the tokens that are about to be pruned or retained. 
\setlength{\textfloatsep}{8pt}
\begin{algorithm}[tb]
\caption{
   Three-step Training Procedure for Learnable Threshold Token Pruning 
   }
\label{alg:lttp}

\begin{algorithmic}
\State {\bfseries Input:} $\mathcal{M}$: model finetuned on target downstream task
\vskip 0.075in
\State \textbf{Step 1:} Apply soft mask to $\mathcal{M}$ and train both the thresholds and model parameters
\hspace*{\fill}{$\triangleright$ Soft Pruning}
\State \textbf{Step 2:} Binarize the mask and fix the thresholds    
\State \textbf{Step 3:} Finetune the model parameters
\hspace*{\fill}{$\triangleright$ Hard Pruning}
\end{algorithmic}
\end{algorithm}
\vspace{5mm}

\begin{figure*}[!t]
\centering
\includegraphics[width=1\textwidth]{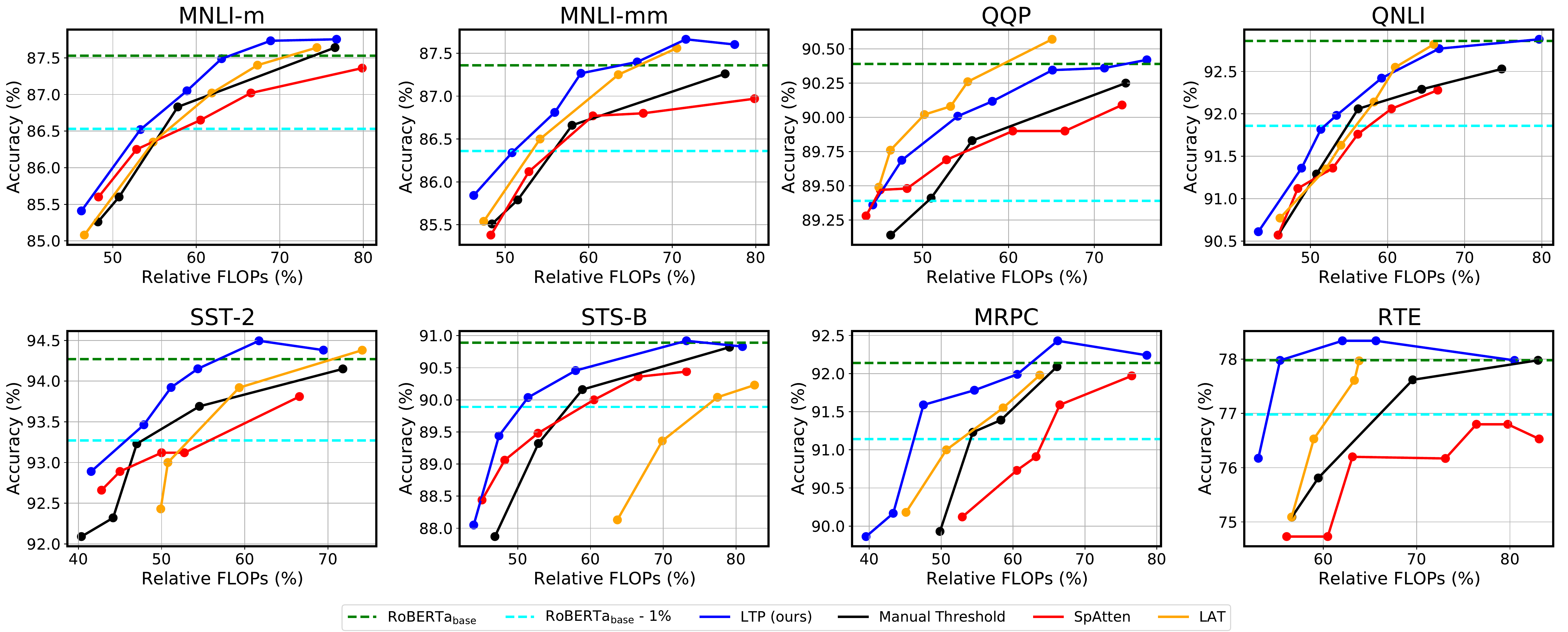}
\caption{Performance of different pruning methods on GLUE tasks for different token pruning methods across different relative FLOPs, i.e., normalized FLOPs with respect to the the baseline model. 
Manual threshold assigns linearly raising threshold values for each layer instead of learning them.
The performance of the baseline model without token pruning (RoBERTa$_\text{base}$) and the model with 1\% performance drop  (RoBERTa$_\text{base}$ - 1\%) are dotted in horizontal lines for comparison. 
}
\label{fig:main_results}
\end{figure*}

\textbf{Regularization.} 
It is not possible to learn $\theta$ without regularization, as the optimizer
generally gets a better loss value if all tokens are present. As such, we add a regularization
term to penalize the network if tokens are left unpruned. This is achieved by imposing
an L1 loss on the masking operator $\tilde{M}$:
\begin{equation}
\mathcal{L}_{\mathrm{new}} = \mathcal{L} + \lambda \mathcal{L}_{\mathrm{reg}} \enspace \textrm{where} \enspace 
\mathcal{L}_{\mathrm{reg}} = \frac{1}{L} \sum_{l=1}^{L} ||\tilde{M}^{(l)}(\text{x})||_{1}.
\label{eq:reg}
\end{equation} 
Here, $\mathcal{L}$ is the original loss function (e.g., cross-entropy loss), and $\lambda$ is the regularization parameter.
Larger values of $\lambda$ result in higher pruning ratios.
This regularization operator induces an additional gradient to the threshold:
\begin{equation}
     \frac{d\mathcal{L}_{\mathrm{reg}}}{d\theta^{(l)}}
     = \frac{1}{d\theta^{(l)}}||\tilde{M}^{(l)}(\text{x})||_{1}
     = \sum_{i=1}^{n} \frac{d\tilde{M}^{(l)}(\text{x}_i)}{d\theta^{(l)}}
\end{equation}
If there are more tokens near the threshold, then the gradient $d \mathcal{L}_{\mathrm{reg}}/d\theta^{(l)}$ is larger.
As a result, the threshold is pushed to a larger value, which prunes more tokens near the threshold boundary.

\vspace{4mm}

\section{\textbf{Experiments}}
\label{sec:results}
\vspace{1mm}

\subsection{Experiment Setup}
\label{sec:experiment_setup}
We implemented \OURS on  RoBERTa$_\text{base}$~\cite{liu2019roberta} using HuggingFace's repo\footnote{\url{https://github.com/huggingface/transformers/}}
and tested on (English) GLUE tasks \cite{wang2018glue} and SQuAD 2.0~\cite{rajpurkar2018know}.
For GLUE tasks \cite{wang2018glue}, we use 6 tasks for evaluation
including sentence similarity (QQP~\cite{iyer2017first}, MRPC~\cite{dolan2005automatically}, STS-B~\cite{cer2017semeval}), 
sentiment classification (SST-2~\cite{socher2013recursive}), 
textual entailment (RTE~\cite{dagan2005pascal}) 
and natural language inference (MNLI~\cite{williams2017broad}, QNLI~\cite{rajpurkar2016squad}).
For evaluating the results, we measure classification accuracy and F1 score for MRPC and QQP, Pearson Correlation and Spearman Correlation for
STS-B, and classification accuracy for the remaining tasks on validation sets.
For the tasks with multiple metrics (i.e., MRPC, QQP, STS-B), we report their average.
For SQuAD 2.0~\cite{rajpurkar2018know}, which is a question and answering task, we measure F1 score for evaluating the results.

As mentioned in~\sref{sec:learned_threshold}, the training procedure of \OURS consists of two stages:
soft pruning that trains both the model parameters and thresholds on downstream tasks, 
followed by hard pruning that fine-tunes the model parameters with fixed thresholds.
We also compare \OURS with the current state-of-the-art token pruning methods of SpAtten~\cite{wang2020spatten} and LAT~\cite{kim2021lat} following the implementation details in their papers.
See~\ref{sec:training_details} for the details of the training process.
We use PyTorch 1.8 throughout all experiments.
For CPU inference speed experiments, we use an Intel Haswell CPU with 3.75GB memory of Google Cloud Platform.
For GPU inference speed experiments, we use an AWS p3.2xlarge instance that has a NVIDIA V100 GPU with CUDA 11.1.

An important issue in previous work~\cite{goyal20powerbert, kim2021lat}  is that \textit{all} input sequences for a specific task are padded to the
nearest power of 2 from the 99th percentile of the sequence lengths, and then the pruned performance is compared with the padded baseline.
This results in exaggerated performance gain over the baseline.
For instance, in~\cite{goyal20powerbert}, inputs from the SST-2 dataset are padded to 64, while its average sentence length is 26 (cf.~\fref{fig:glue-distributions}).
With this approach, one can achieve roughly $2.5\times$ speedup by just removing padding. 
As such, we avoid any extra padding of input sequences and all speedups and throughputs we report are compared with the unpadded
baselines.

\setlength{\textfloatsep}{20pt}
\begin{table}[!t]
\caption{ 
Detailed performance and efficiency comparison of \OURS applied to RoBERTa$_\text{base}$. 
}
\vspace*{-2mm}
\label{tab:main_result}
\centering
\centerline{

{
        \setlength{\tabcolsep}{4pt}{
      \begin{tabular}[t]{c|c>{\columncolor{orange!10}}c|c>{\columncolor{orange!10}}c|>{\columncolor{orange!10}}c}
        \toprule
        \ha  \multirow{2}{*}{Task}  & \multicolumn{2}{c|}{Accuracy Metric} & \multicolumn{2}{c|}{GFLOPs} & \multicolumn{1}{c}{Speedup} \\ 
        \rule{0pt}{1.02\normalbaselineskip}
             & RoBERTa  & \OURS & RoBERTa  & \OURS & \OURS \\
        \hline \rule{0pt}{1.1\normalbaselineskip}
        MNLI-m & 87.53 & 86.53 & 6.83 & 3.64 & 1.88$\times$ \\
        MNLI-mm & 87.36 & 86.37 & 7.15 & 3.63 & 1.97$\times$ \\
        QQP & 90.39 & 89.69 & 5.31 & 2.53 & 2.10$\times$\\
        QNLI & 92.86 & 91.98 & 8.94 & 4.77 & 1.87$\times$ \\
        SST-2 & 94.27 & 93.46 & 4.45 & 2.13 & 2.09$\times$ \\
        STS-B & 90.89 & 90.03 & 5.53 & 2.84 & 1.95$\times$ \\
        MRPC & 92.14 & 91.59 & 9.33 & 4.44 & 2.10$\times$ \\
        RTE & 77.98 & 77.98 & 11.38 & 6.30 & 1.81$\times$ \\
        SQuAD 2.0 & 83.04 & 82.25 & 32.12 & 16.99 & 1.89$\times$ \\
        
        \bottomrule
        \end{tabular} 
        }
        }
}
\end{table}

\begin{table}[!t]
\caption{  
Quantiles (Q1/Q2/Q3) and KL divergence of sentence lengths of training and evaluation datasets for GLUE tasks.
KL divergence are measured after binning the sentence lengths into 20 bins for RTE, MRPC, and STS-B and 50 bins for the others. 
}
 \vskip -2mm
\label{tab:glue_statistics}
    \centerline{
    \centering
    \centerline{
    {
    \setlength{\tabcolsep}{4.5pt}{
       \begin{tabular}{c|cc|c}
        \toprule
        \ha  Task & Quantiles (train) & Quantiles (eval) &  KL Div. \\ 
        \midrule         
        \ha MNLI-m &  27/38/50 & 26/37/50 & 0.0055 \\   
        \ha MNLI-mm & 27/38/50 & 29/39/51 & 0.0042 \\
        \ha QQP & 23/28/36 & 23/28/36 & 0.0006\\
        \ha QNLI & 39/48/59 & 39/49/61 & 0.0092\\
        \ha SST-2 & 7/11/19 & 18/25/33 & 1.2250\\
        \ha STS-B & 20/24/32 & 21/29/41 & 0.0925 \\
        \ha MRPC & 45/54/63 & 45/54/64 & 0.0033\\
        \ha RTE & 44/57/86 & 42/54/78 & 0.0261 \\
        
        \bottomrule
        \end{tabular} 
       
        }
        }
    }
}
\vspace{-1mm}
\end{table}

\subsection{Performance Evaluation}
\label{sec:evaluation}
Table~\ref{tab:main_result} lists the accuracy and GFLOPs for \OURS.
We select a model for each downstream task that achieves the smallest GFLOPs while
constraining the accuracy degradation from the baseline (RoBERTa$_\text{base}$) to be at most 1\%.
Using our method, sequence lengths in each layer can vary across different input sentences.
Therefore, we report the averaged GFLOPs of processing all input sentences in the development set. 
As shown in the table, our method achieves speedup of 1.96$\times$ on average and up to 2.10$\times$ within 1\% accuracy degradation.

Figure~\ref{fig:main_results} plots the accuracy of \OURS (blue lines) as well as the prior pruning methods (red lines for SpAtten and orange lines for LAT) with different FLOPs on GLUE tasks.
\OURS consistently outperforms SpAtten for all tasks with up to \textasciitilde2\% higher accuracy under the same amount of FLOPs.
Compared with LAT, \OURS outperforms for all tasks except for QQP  with up to \textasciitilde2.5\%  higher accuracy for the same target FLOPs.
For QQP alone, \OURS achieves at most \textasciitilde0.2\% lower accuracy than LTP.

An important observation is that for SST-2 and STS-B where LTP (ours) outperforms LAT with large margins, the sequence length varies greatly from the training dataset to the evaluation dataset 
as can be seen from the large KL-divergence in~\tref{tab:glue_statistics} and \fref{fig:glue-distributions} (b, c).
On the other hand, for QQP, the only dataset that LAT slightly outperforms LTP (ours),
the sequence length distribution of the training dataset is  almost identical to that of the evaluation dataset as can be seen from the small KL-divergence in~\tref{tab:glue_statistics} and  \fref{fig:sentence-examples} (a). 
This observation supports our claim in~\sref{sec:intro} and~\ref{sec:relatedwork}:
\OURS is robust to sequence length variations as it does not fix the pruning configuration unlike other methods using a single pruning configuration regardless of the input sequence length.
This is important in practice because the sequence lengths during inference do not always follow the sequence length distribution of the training dataset as in SST-2 and STS-B. 
We make a further discussion in~\sref{sec:robustness_eval}.

\begin{figure}[!t]
\centering
  \includegraphics[width=0.48\textwidth]{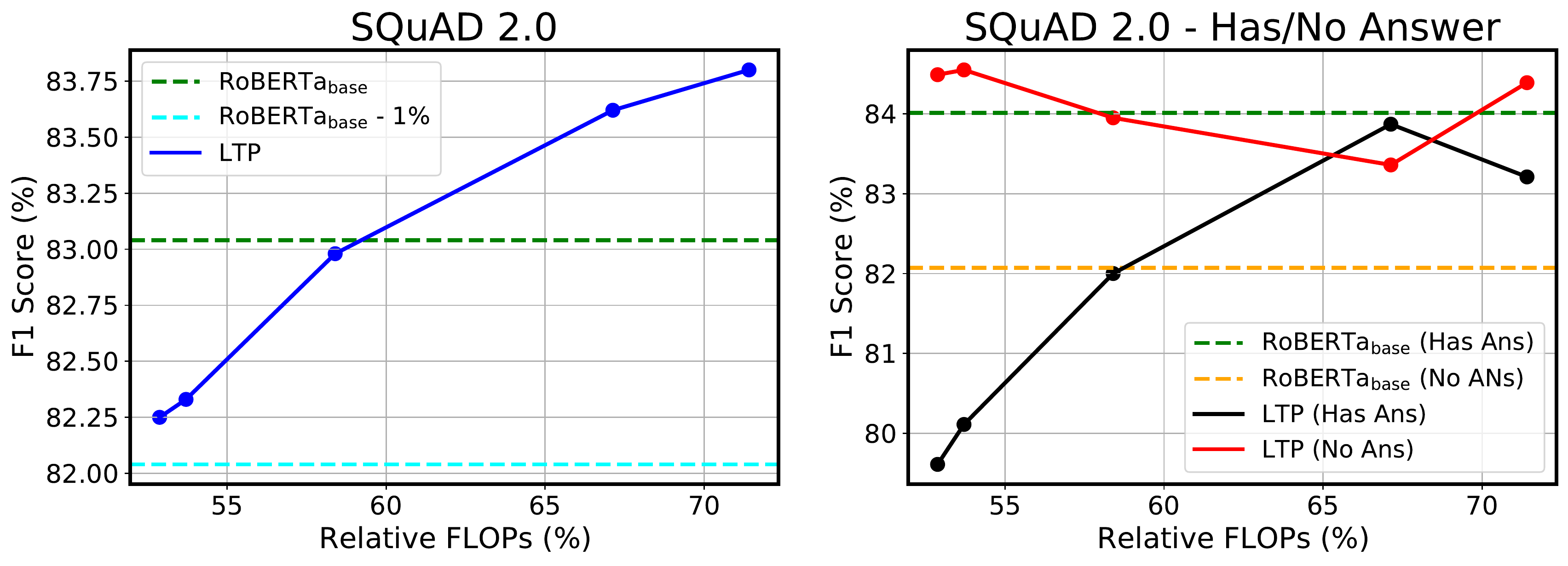}
  \caption{
  (Left) Performance of LTP on SQuAD 2.0 across different relative FLOPs with respect to the the baseline model on the full validation set.
  (Right) Performance of LTP on the subsets of the validation set, 
  one with answers (Has Ans, black) and the other without (No Ans, red).
  The performance of the baseline model without token pruning (RoBERTa$_\text{base}$) and the model with 1\% performance drop  (RoBERTa$_\text{base}$ - 1\%) are dotted in horizontal lines for comparison. 
  }
  \label{fig:squad}
\end{figure}

For SQuAD 2.0, we have similar results to GLUE. 
As can be seen in \tref{tab:main_result} and \fref{fig:squad} (Left),
we obtain nearly identical F1 score to baseline at 0.58 relative FLOPs,
and 1.89$\times$ speedup with less than 1\% drop of F1 score.
The SQuAD 2.0 dataset is divided into two subsets: the subset of examples where the answer to the question is included in the context text, and the subset that has no answer.
In \fref{fig:squad} (Right), we further plot the results on each subset of the dataset (black and red for the one with and without answers, respectively).
We see that the F1 score decreases for the subset with answers and increases for the subset without answers as we decrease the relative FLOPs.
This is to be expected as the question answering head will predict no answer if the start and end points of the answer within the context cannot be determined due to high token pruning ratios.
Thus, a careful setting of $\lambda$ in \eref{eq:reg} is necessary to balance the accuracy between the two subsets.

At last, we also highlight that
\OURS has an additional gain over the prior \topk based approaches by avoiding computationally inefficient \topk operations as further discussed in~\sref{sec:computation_efficiency}.


\begin{figure*}[!t]
\centering
\includegraphics[width=1\textwidth]{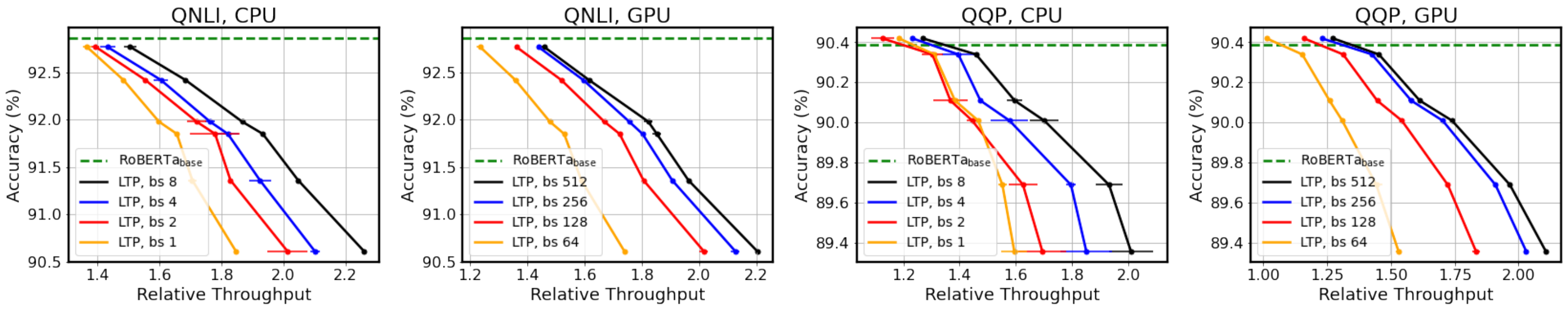}
\caption{Relative throughput of \OURS with respect to the baseline without token pruning (RoBERTa$_{\text{base}}$) with different batch sizes on Intel Haswell CPU and NVIDIA V100 GPU. 
The performance of RoBERTa$_{\text{base}}$ are dotted in horizontal lines.
}
\label{fig:throughput}
\end{figure*}

\subsection{Robustness to Sequence Length Variation}
\label{sec:robustness_eval}

In~\sref{sec:evaluation}, we claim that \OURS is more robust against sequence length variations from training time to evaluation time.
Here, we make a more systematic analysis on this.
Ideally, performance should be independent of sequence length.
To quantitatively test the robustness of pruning methods against sequence length variations, we train \OURS and LAT on QNLI and QQP,
but only using the training examples whose sequence lengths are below the median length of the evaluation dataset.
We then evaluate the resulting models using the evaluation examples with sequence lengths (i) below the median (\textasciitilde Q2), (ii) between the median and the third quantile (Q2\textasciitilde Q3), and (iii) above the third quantile (Q3\textasciitilde) of the evaluation dataset.
To make a fair comparison, we choose models from \OURS and LAT that require similar FLOPs on \textasciitilde Q2.

The results are listed in~\tref{tab:robustness_length}. \OURS consistently achieves better accuracy and FLOPs over different sequence lengths, even with those that are significantly longer than the training sequences.
On the contrary, LAT shows significant accuracy degradation as longer sequences are over-pruned, which can be seen from the significant FLOPs reduction.
In particular, \OURS outperforms LAT by up to 16.44\% and 9.20\% on QNLI and QQP for the Q3$\text{\textasciitilde}$ evaluation dataset.

\begin{table}[!t]
\caption{ 
LTP and LAT trained with the sequences shorter than the median length,
and evaluated with the sequences shorther than the median (\textasciitilde Q2), between the median and the third quantile (Q2\textasciitilde Q3), and longer than the third quantile (Q3\textasciitilde) of the evaluation dataset.
FLOPs are relative FLOPs (\%) with respect to RoBERTa$_{base}$.
}
\vspace*{-2mm}
\label{tab:robustness_length}
\centering
\centerline{

{
        \setlength{\tabcolsep}{3pt}{
      \begin{tabular}[t]{c|c|ccc|ccc}
        \toprule
        \ha  \multirow{2}{*}{}  & \multirow{2}{*}{Task} & \multicolumn{3}{c|}{QNLI} & \multicolumn{3}{c}{QQP} \\ 
        \  &   & \textasciitilde Q2 & Q2\textasciitilde Q3 & Q3\textasciitilde
               & \textasciitilde Q2 & Q2\textasciitilde Q3 & Q3\textasciitilde \\
        \midrule
        \ha \OURS   & Acc. & 91.21 & 90.02 & 91.81 & 89.42 & 89.51 & 91.37 \\
        \ha  (ours)   & FLOPs & 55.89 & 55.60 & 56.02  & 55.18 & 56.29 & 58.01 \\
        \midrule
        \ha LAT & Acc. & 90.87 & 86.12 & 75.37 & 89.20 & 87.27 & 82.17   \\
        \ha     & FLOPs & 56.21 & 46.55 & 35.89 & 55.17 & 46.61 & 34.14 \\
        \midrule
        \hc Diff.   & Acc. & +0.34 & +3.90 & +16.44 & +0.22 & +2.24 & +9.20\\
        \bottomrule
        \end{tabular} 
        }
        }
}
\vspace{-5mm}
\end{table}

\subsection{Ablation Studies}
\label{sec:ablation_studies}
Instead of learning thresholds, we can set them manually.
Because manually searching over the exponential search space is intractable, we add a constraint to the search space by assigning linearly rising threshold values for each layer, similar to how SpAtten~\cite{wang2020spatten} assigns the token retain ratios:
given the threshold of the final layer $\theta^{(L)}$, the threshold for layer $l$ is set as $\theta^{(L)}l/{L}$.
We plot the accuracy and FLOPs of the manual threshold approach in~\fref{fig:main_results} as black lines.
While this approach exhibits decent results on all downstream tasks, the learned thresholds consistently outperform the manual thresholds under the same FLOPs.
This provides empirical evidence for the effectiveness of our threshold learning method.

\subsection{Direct Throughput Measurement on Hardware}
\label{sec:throughput}
We directly measure throughputs on real hardware by deploying \OURS on a NVIDIA V100 GPU and a Intel Haswell CPU.
For inference, we completely remove the pruned tokens and rearrange the retained tokens into a \textit{blank-free} sequence to have a latency gain. 
One consequence of adaptive pruning, however, is that each sequence will end up with a different pruning pattern and sequence length.
As such, a naive hardware implementation of batched inference may require padding all the sequences in a batch to ensure that they all have the same length (i.e., the maximum sequence length in the batch),
which results in a significant portion of computation being wasted to process padding tokens.
To avoid this, we use NVIDIA's Faster Transformer\footnote{\url{https://github.com/NVIDIA/FasterTransformer}}
for GPU implementation that requires large batch sizes.
This framework dynamically removes and inserts padding tokens during inference so that most of the transformer operations effectively skip processing padding tokens.
This enables fast inference even with irregular pruning lengths of individual sequences.
For the CPU implementation, we find naive batching (i.e., padding sequences to the maximum sentence length) enough for good throughput.

The measured throughput results are shown in~\fref{fig:throughput} for different batch sizes.
For all experiments, relative throughput is evaluated 3 times on the randomly shuffled datasets.
\OURS achieves up to $\sim$1.9$\times$ and $\sim$2.0$\times$ thoughput improvement for QNLI and QQP on both CPU and GPU, as compared to the baseline.
This is similar to the theoretical speedup inferred from the  FLOPs reduction reported in~\tref{tab:main_result}.
Importantly, the speedup of \OURS increases with larger batch sizes on both CPU and GPU, proving effectiveness of \OURS on batched cases.

\subsection{\OURS with Quantization and Knowledge Distillation}
\label{sec:quantization_results}


\begin{figure}[!t]
\centering
  \includegraphics[width=0.48\textwidth]{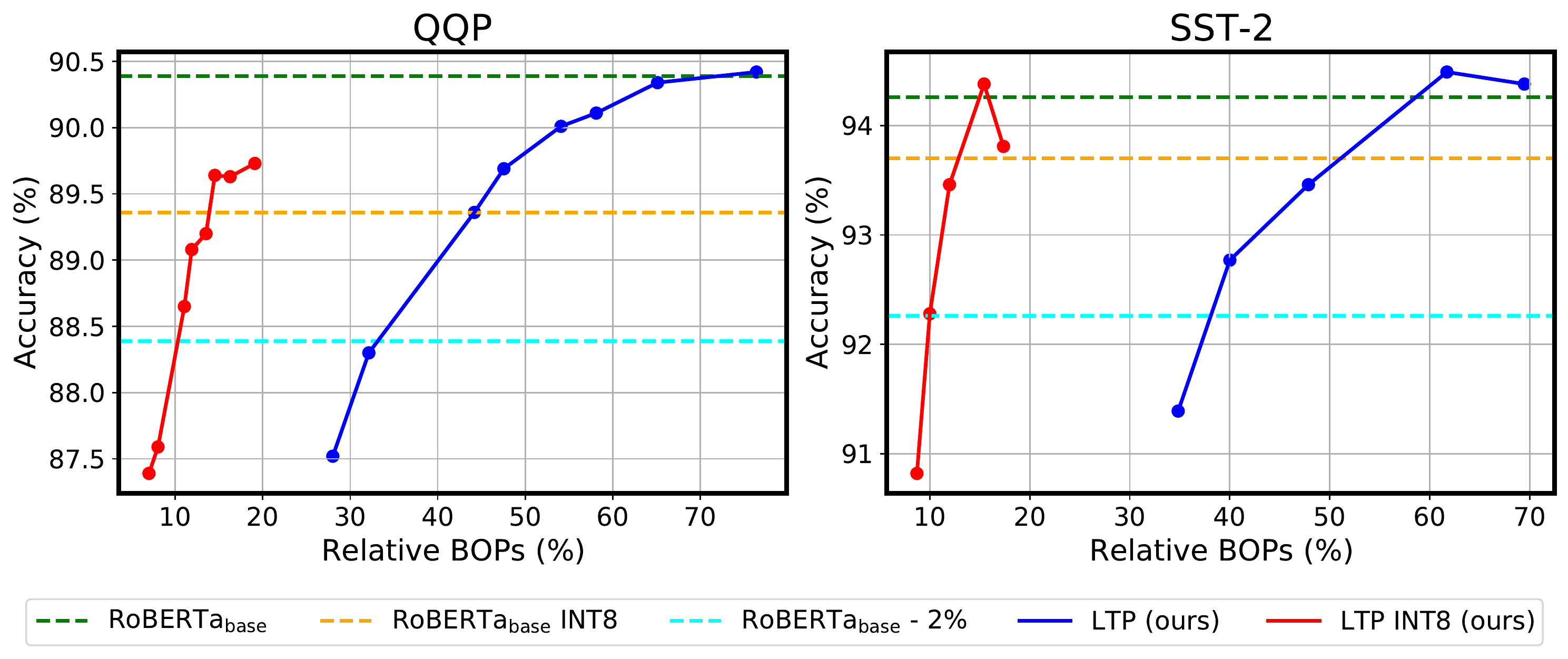}
  \caption{
  Accuracy and relative BOPs of the FP16 baselines and INT8 \OURS models on QQP and SST-2 datasets.
  Note that FP16 unpruned RoBERTa$_\text{base}$ is used as the baseline.
  Thus, INT8 quantization of the models translates to $4\times$ reduction in relative BOPs.
  }
  \label{fig:int8}
  \vspace{-4mm}
\end{figure}


Here, we show that our token-level pruning method is compatible with other compression methods.
In particular, we perform compression experiments by combining \OURS with quantization and knowledge distillation~\cite{hinton2015distilling} together.
For quantization,
we use the static uniform symmetric integer quantization method~\cite{gholami2021survey}, which is easy to deploy in commodity hardware with minimal run-time overhead.
All the model parameters are quantized to 8-bit integers, except for those of the embedding layer whose bit-width does not affect the inference speed.
Afterwards, we apply knowledge distillation that helps recover accuracy for high compression ratios.
We set the baseline RoBERTa$_\text{base}$ model as the teacher and the quantized \OURS model as the student.
We then distill knowledge from the teacher model into the student model through a knowledge distillation loss 
that matches the output logits of the classification layer and the output representations of the embedding layer in the teacher model 
to the counterparts in the student model.
The training objective is a convex combination of the original loss and the knowledge distillation loss.
As shown in~\fref{fig:int8}, we achieve up to 10$\times$ reduction in bit operations (BOPs) with less than $2\%$ accuracy degradation as compared to FP16 RoBERTa$_\text{base}$ by combining quantization and knowledge distillation.
The results empirically show the effectiveness of \OURS with other compression methods.

\section{\textbf{Conclusions}}
\label{sec:conclusions}

In this work, we present Learned Token Pruning (\OURS), a fully automated token pruning framework for transformers.
\OURS only requires comparison of token importance scores with threshold values to determine unimportant tokens, thus adding minimal complexity over the original transformer inference.
Importantly, the threshold values are learned for each layer during training through a differentiable soft binarized mask that enables backpropagation of gradients to the threshold values.
Compared to the state-of-the-art token pruning methods, \OURS outperforms by up to \textasciitilde2.5\% accuracy with the same amount of FLOPs.
Extensive experiments on GLUE and SQuAD show the effectiveness of \OURS, as it achieves up to 2.10$\times$ FLOPs reduction over the baseline model within only 1\% of accuracy degradation.
Our preliminary (and not highly optimized) implementation shows up to 1.9$\times$ and 2.0$\times$ throughput improvement on an Intel Haswell CPU and a NVIDIA V100 GPU.
Furthermore, \OURS exhibits significantly better robustness and consistency over different input sequence lengths.

\vspace{3mm}



\bibliographystyle{ACM-Reference-Format2}
\bibliography{_main}
\clearpage
\appendix
\counterwithin{figure}{section}
\counterwithin{table}{section}

\section{Appendix} 
\subsection{Training Details}

\label{sec:training_details}
The training procedure of \OURS consists of two separate stages: soft pruning followed by hard pruning.
For soft pruning, we train both the model parameters and the thresholds on downstream tasks for 1 to 10 epochs, depending on the dataset size.
We find it effective to initialize the thresholds with linearly rising values as described in~\ref{sec:ablation_studies} with a fixed threshold of the final layer.
We search the optimal temperature $T$ in a search space of \{{1, 2, 5, 10, 20}\}e-4
and vary $\lambda$ from 0.001 to 0.4 to control
the number of tokens to be pruned (and thus the FLOPs) for all experiments. 
We then fix the thresholds and perform an additional training with the hard pruning to fine-tune the model parameters only.
More detailed hyperparameter settings are listed in~\tref{tab:hyperparam} for GLUE and SQuAD 2.0.

SpAtten is trained based on the implementation details in the paper:
the first three layers retain all tokens and the remaining layers are assigned with linearly decaying token retain ratio
until it reaches the final token retain ratio at the last layer.
We vary the final token retain ratio from 1.0 to -1.0 (prune all tokens for non-positive retain ratios) to control the FLOPs of SpAtten. 
For both \OURS and SpAtten, we use learning rate of \{0.5, 1, 2\}e-5, except for the soft pruning stage of \OURS where we use 2e-5. 
We follow the optimizer setting in RoBERTa~\cite{liu2019roberta} and use batch size of 64 for all experiments.

LAT is trained using the same hyperparameter and optimizer setting in the paper except for the length drop probabilities:
for more extensive search on more aggressive pruning configurations, we used 0.25, 0.3, 0.35, and 0.4 for the length drop probability instead of 0.2 in the original setting.

\begin{figure*}[!t]
\centering
  \includegraphics[width=0.95\textwidth]{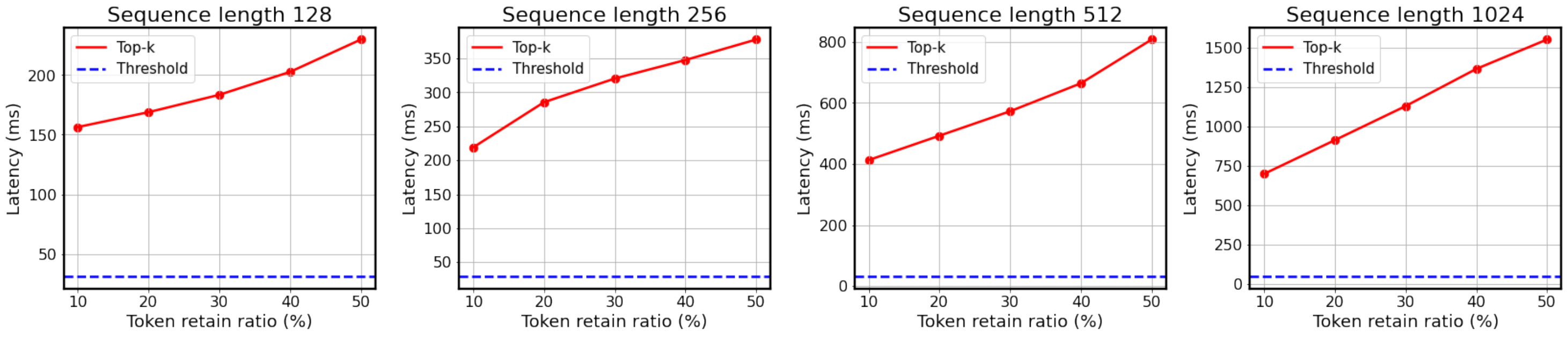}
  \caption{Wall-clock latency comparison between \topk operation and threshold operation on an Intel Haswell CPU for different sequence length across various token retain ratios. 
  Note that the latency of a threshold operation is independent of sequence length.
  }
  \label{fig:time_perf}
  \vspace{-4mm}
\end{figure*}
\subsection{Computation Efficiency Comparison}
\label{sec:computation_efficiency}
Here we compare the efficiency of \topk versus threshold operation.
To do this, we use
a batch size of 32 and average the latency over 1000 independent runs.
For each sequence length, we test over five different token retain ratios from 10\% to 50\%
(e.g., 10\% token retain ratio is the case where we select \topk 10\% of tokens from the input sequence).

With the above setting, we directly measure the latency of these two
operations on an Intel Haswell CPU, and report the results in~\fref{fig:time_perf}. 
For \topk operation, there is a noticeable increase in latency when token retain ratios and sequence lengths become larger 
whereas this is not an issue for our threshold pruning method as it only requires a comparison operation.
More importantly, \topk operation incurs a huge latency overhead that is up to 7.4$\times$ and 33.4$\times$ slower than threshold operation for sequence length of 128 and 1024, respectively.\footnote{
The inefficiency of \topk is also further confirmed by~\cite{wang2020spatten}, where they report only 1.1$\times$ speedup for GPT-2
without the \topk hardware engine that they developed.
}

\begin{table}[!t]
\caption{  
Detailed hyperparameters for \OURS training.
}
 \vskip -2mm
\label{tab:hyperparam}
    \centerline{
    \centering
    \centerline{
    {
    \setlength{\tabcolsep}{4.5pt}{
       \begin{tabular}{c|c|cc}
        \toprule
        \ha  Stage & Hyperparam   & GLUE & SQuAD 2.0 \\ 
        \midrule         
        \ha \multirow{5}{*}{\shortstack{Soft \\ pruning}} & epochs & 1 - 10 &  1\\
                  & learning rate & 2e-5 & 2e-5 \\
                  & $T$ & $\{$1, 2, 5, 10, 20$\}$e$\text{-}$4 & $\{$1, 10$\}$e-4 \\
                  & $\lambda$ & 0.001 - 0.2 & 0.001 - 0.4 \\
                  & init. final thres. & 0.01 & 0.003 \\
        \midrule
        \ha Hard    & epochs & 10 & 5\\
        \ha pruning & lr & \{0.5, 1, 2\}e-5 & \{0.5, 1, 2\}e-5 \\
        \bottomrule
        \end{tabular} 
       
        }
        }
    }
}
\end{table}


\subsection{Discussion}
\label{sec:appendix_discussion}

\begin{figure*}[!t]
\centering
\subfloat[SST-2]{\includegraphics[width=.42 \textwidth]{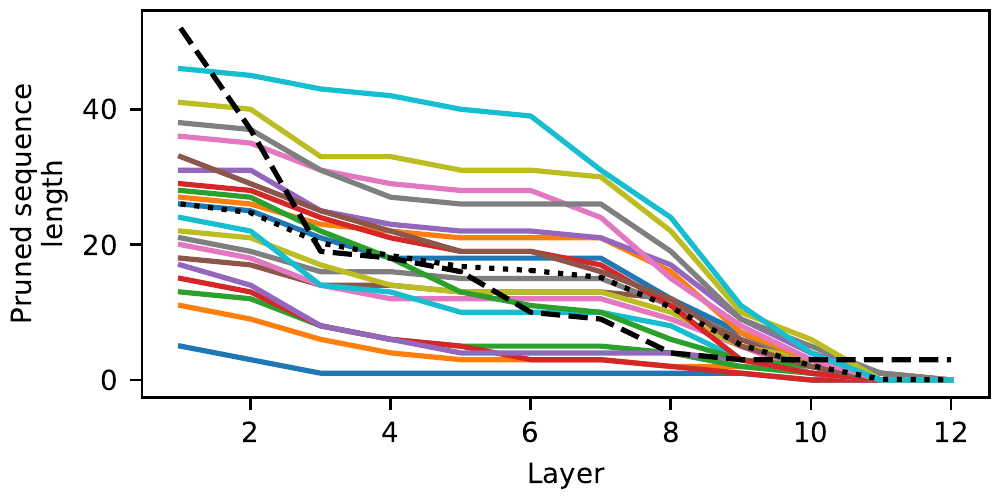}}\qquad
\subfloat[MNLI-m]{\includegraphics[width=.42 \textwidth]{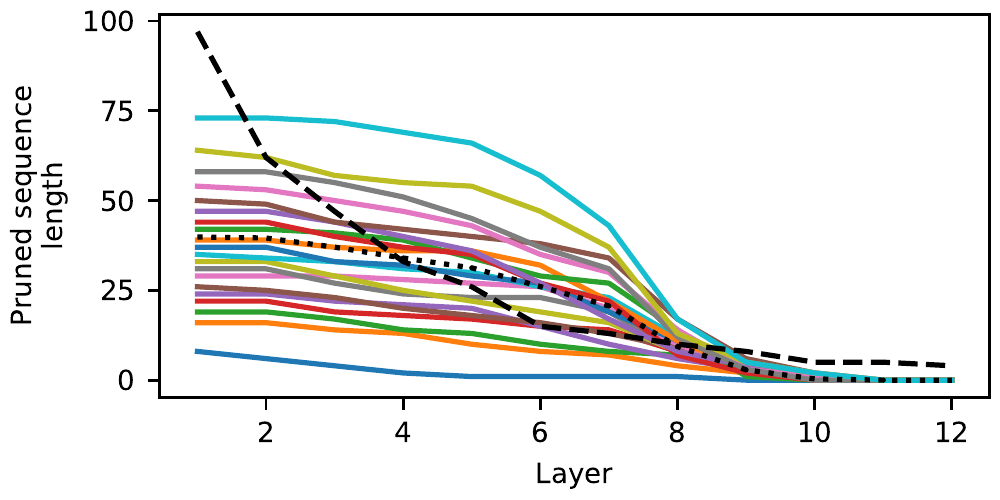}}\\
  \caption{
  Sample trajectories of pruned sequence length as the sequences are passed through model layers. For \OURS, 20 samples were evenly selected from the sets after sorting by initial sequence length. (a) SST-2. (b) MNLI-m. The mean sequence length for \OURS is shown by a black dotted line, and the LAT baseline is shown by a black dashed line. Parameters were selected so as to provide a 1\% drop in accuracy from baseline for both methods.}
  \label{fig:sentence-lengths-throughout}
\end{figure*}

\begin{figure*}
\centering
\subfloat[SST2]{\includegraphics[width=0.9\textwidth]{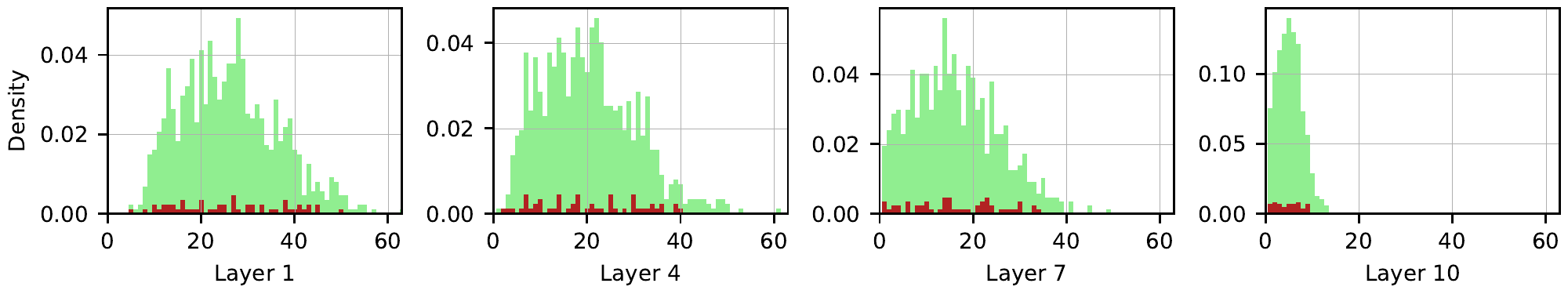}}\\
\vspace{-2mm}
\subfloat[MNLI-m]{\includegraphics[width=0.9\textwidth]{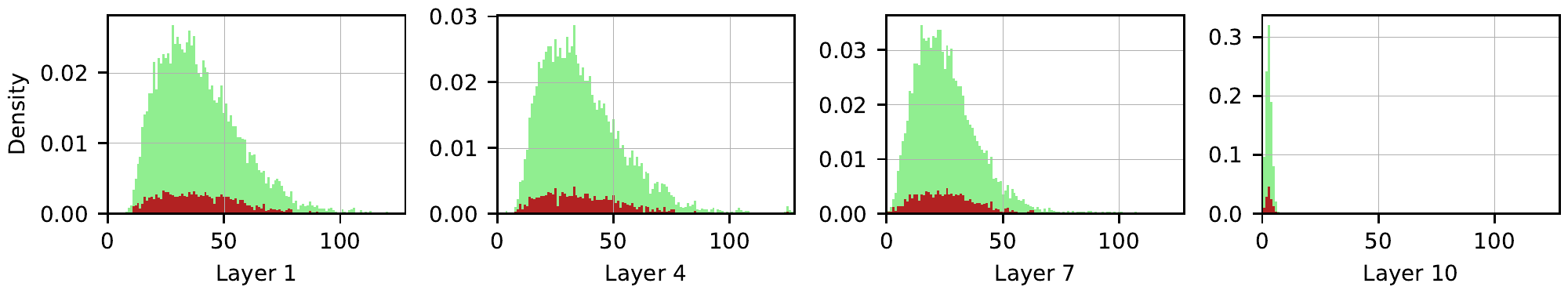}}\\
\caption{
  Histogram of pruned sequence length (x-axis) as the input sequence is processed through different transformer blocks. 
  y-axis shows the relative count of sentences with the particular sequence length in x-axis. 
Green denotes input sequences that are correctly classified, and red denotes incorrect classifications.}
\label{fig:sentence-lengths-accuracy}
\end{figure*}

\subsubsection{Example Sequence Length Trajectories}
\fref{fig:sentence-lengths-throughout} shows how the pruned sequence length decreases for input sequences of varying lengths. 
For LAT, the token pruning configuration is fixed for all sequences in the dataset. In \OURS, token pruning can be more or less aggressive depending on the sequence content and the number of important tokens in the sequence. 
On average, \OURS calculates 25.86\% fewer tokens per layer than LAT for MNLI-m and 12.08\% fewer tokens for SST-2.
For both \OURS and LAT, the model has been trained to produce a 1\% drop in accuracy compared to baseline.

\vspace{2mm}
\subsubsection{Unbiased Token Pruning for Various Sequence Length}
Figure \ref{fig:sentence-lengths-accuracy} shows the distributions of initial sequence lengths
for sequences that are correctly classified and for sequences that are not. We see that
for multiple tasks, there is no significant correlation between the length of
the sequence and the accuracy of the pruned models. 
Importantly, this suggests that our method is not biased towards being more accurate on longer or shorter sequences.

\vspace{2mm}

\subsection{Comparison with TR-BERT on GLUE}
\label{sec:tr-bert comparison}
Unlike LAT and SpAtten, TR-BERT \cite{ye2021trbert} does not report results on the GLUE benchmark tasks described in the paper. We attempted to run TR-BERT on the GLUE tasks using the TR-BERT repo\footnote{\url{https://github.com/thunlp/TR-BERT}}, but were unable to get the algorithm to converge to a high accuracy, despite varying the learning rate between 1e-6 and 1e-3 and the value of $\alpha$, the parameter that defines the length penalty, over the search space of $\{0.01, 0.05, 0.1, 0.5, 1, 2, 5\}$. We also varied the number of training epochs based on the number of examples in each task's training set. The authors of TR-BERT note the convergence difficulties of RL learning while describing the algorithm in their paper.

\end{document}